\documentclass[10pt,twocolumn,letterpaper]{article}

\usepackage{iccv}              

\definecolor{iccvblue}{rgb}{0.21,0.49,0.74}
\usepackage[pagebackref,breaklinks,colorlinks,allcolors=iccvblue]{hyperref}

\def\paperID{1299} 
\def\confName{ICCV}
\def\confYear{2025}

\title{MaterialMVP: Illumination-Invariant Material Generation \\via Multi-view PBR Diffusion}

\author{
Zebin He\textsuperscript{1,2,3} \quad
Mingxin Yang\textsuperscript{2} \quad
Shuhui Yang\textsuperscript{2} \quad
Yixuan Tang\textsuperscript{2} \quad
Tao Wang\textsuperscript{3} \quad
Kaihao Zhang\textsuperscript{4} \\
Guanying Chen\textsuperscript{1} \quad
Yuhong Liu\textsuperscript{2} \quad
Jie Jiang\textsuperscript{2} \quad
Chunchao Guo\textsuperscript{2$\dagger$} \quad
Wenhan Luo\textsuperscript{3$\ddagger$}
\\[0.5em]
\textsuperscript{1}Shenzhen Campus of Sun Yat-sen University \quad
\textsuperscript{2}Tencent Hunyuan \\
\textsuperscript{3}The Hong Kong University of Science and Technology \\
\textsuperscript{4}Harbin Institute of Technology (Shenzhen)
\\[0.5em]
\texttt{\small hezb5@mail2.sysu.edu.cn} \quad
\texttt{\small \{taowangzj,super.khzhang,guanying2018,whluo.china\}@gmail.com}\\
\texttt{\small \{maddoxyang,oakyang,setskytang,ehomeliu,zeus,chunchaoguo\}@tencent.com}\\
}

\begin{document}


\twocolumn[{
\renewcommand\twocolumn[1][]{#1}
\maketitle
\vspace{-9mm}

\begin{center}
    \captionsetup{type=figure}
    \includegraphics[width=0.97\textwidth]{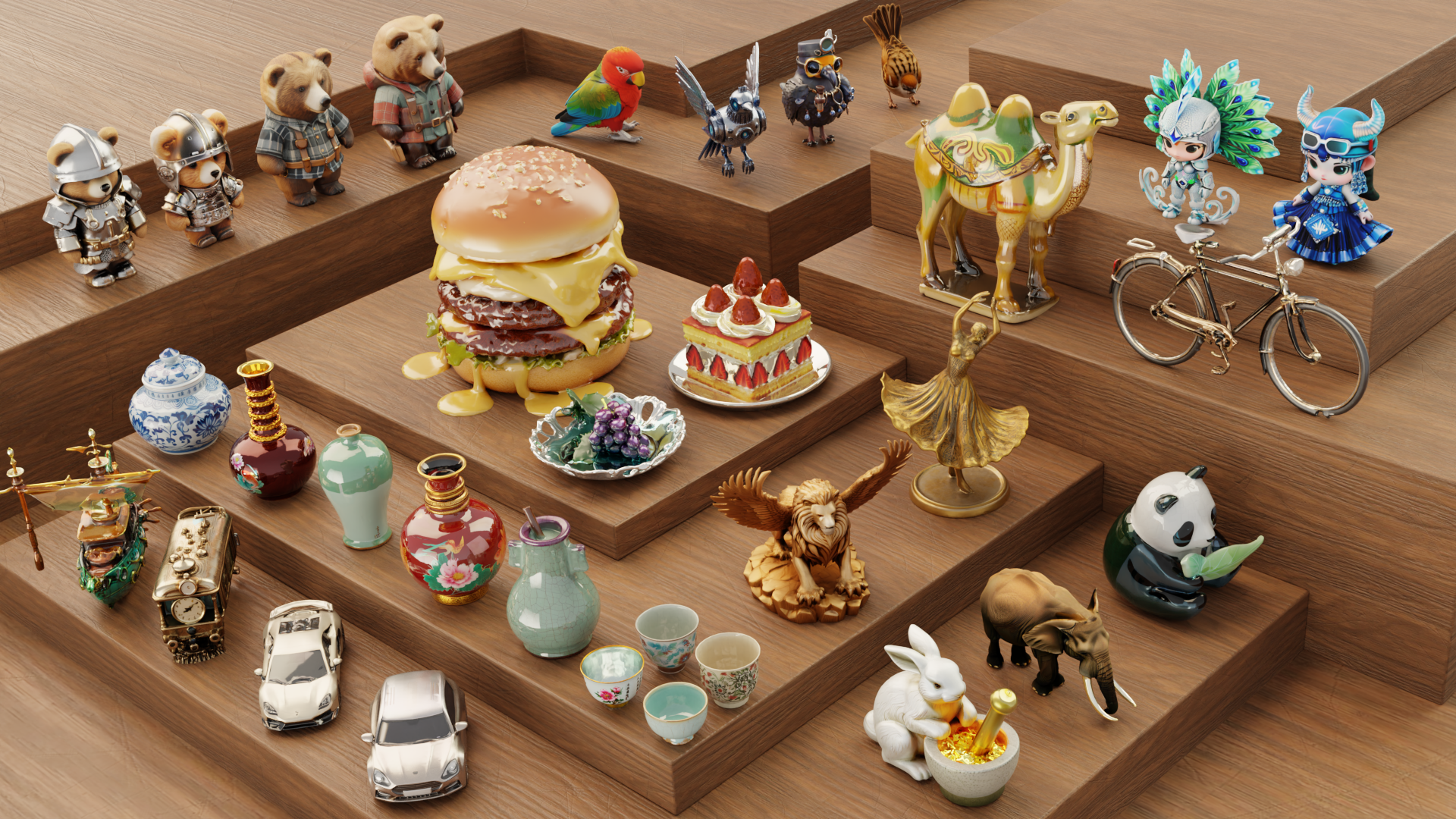}
    \caption{A series of textured meshes with PBR material generated by MaterialMVP, showcasing exceptional texture generation and realistic physical behavior under environmental illumination. More results and code can be found on \url{https://github.com/ZebinHe/MaterialMVP}.}
    \label{fig:teaser}
\end{center}
}]

\def\thefootnote{}\footnotetext{$\dagger$ Project leader.}
\def\thefootnote{}\footnotetext{$\ddagger$ Corresponding author.}

\begin{abstract}
Physically-based rendering (PBR) has become a cornerstone in modern computer graphics, enabling realistic material representation and lighting interactions in 3D scenes. In this paper, we present MaterialMVP, a novel end-to-end model for generating PBR textures from 3D meshes and image prompts, addressing key challenges in multi-view material synthesis. Our approach leverages Reference Attention to extract and encode informative latent from the input reference images, enabling intuitive and controllable texture generation. We also introduce a Consistency-Regularized Training strategy to enforce stability across varying viewpoints and illumination conditions, ensuring illumination-invariant and geometrically consistent results. Additionally, we propose Dual-Channel Material Generation, which separately optimizes albedo and metallic-roughness (MR) textures while maintaining precise spatial alignment with the input images through Multi-Channel Aligned Attention. Learnable material embeddings are further integrated to capture the distinct properties of albedo and MR. Experimental results demonstrate that our model generates PBR textures with realistic behavior across diverse lighting scenarios, outperforming existing methods in both consistency and quality for scalable 3D asset creation.
\end{abstract}    
\section{Introduction}
\label{sec:intro}
Generating textures for 3D models is a crucial task in creating visually compelling and realistic digital assets, with numerous studies dedicated to advancing this field \cite{yu2023texture, richardson2023texture,cao2023texfusion,chen2023text2tex,le2023euclidreamer,tang2024intex,chen2024scenetex,zeng2024paint3d,zhang2024repaint123,bensadoun2024meta,lu2024direct2,zhao2025hunyuan3d,xiang2024make,ji2024reframe,jiang2025flexitex}. While traditional RGB textures focus on color representation, Physically Based Rendering (PBR) textures take a step further by simulating real-world material properties such as reflectivity, roughness, and light interaction, resulting in more immersive and physically plausible visual experiences.

Some methods tackle the task of PBR texture generation through an optimization-based approach guided by text \cite{chen2023fantasia3d, zhang2024dreammat, xu2023matlaber, youwang2024paint, liu2024unidream}. These techniques typically rely on pre-trained 2D diffusion models to provide image priors, employing mechanisms like score distillation sampling \cite{poole2022dreamfusion}. By iteratively optimizing the textured mesh to align with the textual description, they achieve high-quality results. However, the iterative nature of this process makes it computationally expensive and time-consuming.

Image-conditioned approaches are often more intuitive and user-friendly, by providing more direct guidance. Texturedreamer \cite{yeh2024texturedreamer} and Hyperdreamer \cite{wu2023hyperdreamer} extend the optimization-based framework to image-conditioned generation by leveraging DreamBooth \cite{ruiz2023dreambooth}, yet they are still bottlenecked by the lengthy inference time. 
On the other hand, generative approaches like SuperMat \cite{hong2024supermat}, RGB$\leftrightarrow$X \cite{zeng2024rgb}, and IntrinsicAnything \cite{chen2024intrinsicanything} utilize diffusion priors to achieve faster generation. 
However, these methods are limited to single-view generation.
TexGen \cite{yu2024texgen} takes the images in the form of UV maps and demonstrates reasonable results, but struggles to handle the backside of the model and the seams correctly. CLAY \cite{zhang2024clay} proposes to generate multi-view PBR textures from 3D models, leveraging IP-Adapter \cite{ye2023ip} to introduce reference images and achieving impressive results, though it often struggles with precise alignment between the generated textures and the input images.

We introduce MaterialMVP, a one-stage model designed to generate PBR textures for 3D meshes. Our model produces multi-view, high-quality PBR textures, including albedo, metallic, and roughness maps. These textures not only highly follow the input reference images but also ensure precise alignment between different texture maps. 

To be specific, we employ a Multiview Diffusion Model framework designed to generate multiple view-consistent PBR maps from image prompts. To simultaneously produce albedo, metallic, and roughness maps, we introduce a Dual-Channel Material Generation framework, which extends the original diffusion model by adding an extra channel to generate metallic-roughness (MR) alongside the existing albedo channel. For the input reference image, we utilize Reference Attention, leveraging a reference branch to extract detailed information from the input. This ensures the generated textures retain fine details and remain faithful to the reference. To address issues such as unwanted lighting artifacts in diffusion-generated outputs and improve the model’s robustness to input viewpoint perturbations, we introduce Consistency-Regularized Training. This strategy trains the model on pairs of reference images with subtle variations in camera pose and lighting but requires the model to produce identical, lighting-invariant results. This forces the model to learn lighting-invariant and accurate PBR maps across diverse conditions. Additionally, to tackle the misalignment problem in PBR material generation tasks, we design a Multi-Channel Aligned Attention module to synchronize information between the albedo and MR channels, ensuring the output materials are well-aligned and do not produce unexpected shadows or artifacts on textured meshes. At the same time, we incorporate Learnable Material Embeddings for each channel, providing additional context to help each channel learn its unique distribution, resulting in artifact-free and coherent textures.

In summary, our contributions are as follows. 1) We propose MaterialMVP, an end-to-end multi-view PBR material generation model that produces high-quality, multi-channel-aligned, and view-consistent textures. 2) We introduce Consistency-Regularized Training to improve model robustness to viewpoint perturbations and effectively disentangle lighting effects from material properties, producing lighting-invariant and accurate PBR maps. 3) We develop a Dual-Channel Material Generation framework, which processes albedo and metallic-roughness maps in separate channels. Leveraging Multi-Channel Aligned Attention and enhanced by Learnable Material Embeddings, it produces high-quality, coherent, and artifact-free textures.
\section{Related Work}
\label{sec:formatting}

\begin{figure*}[t]
    \centering
    \includegraphics[width=0.9\linewidth]{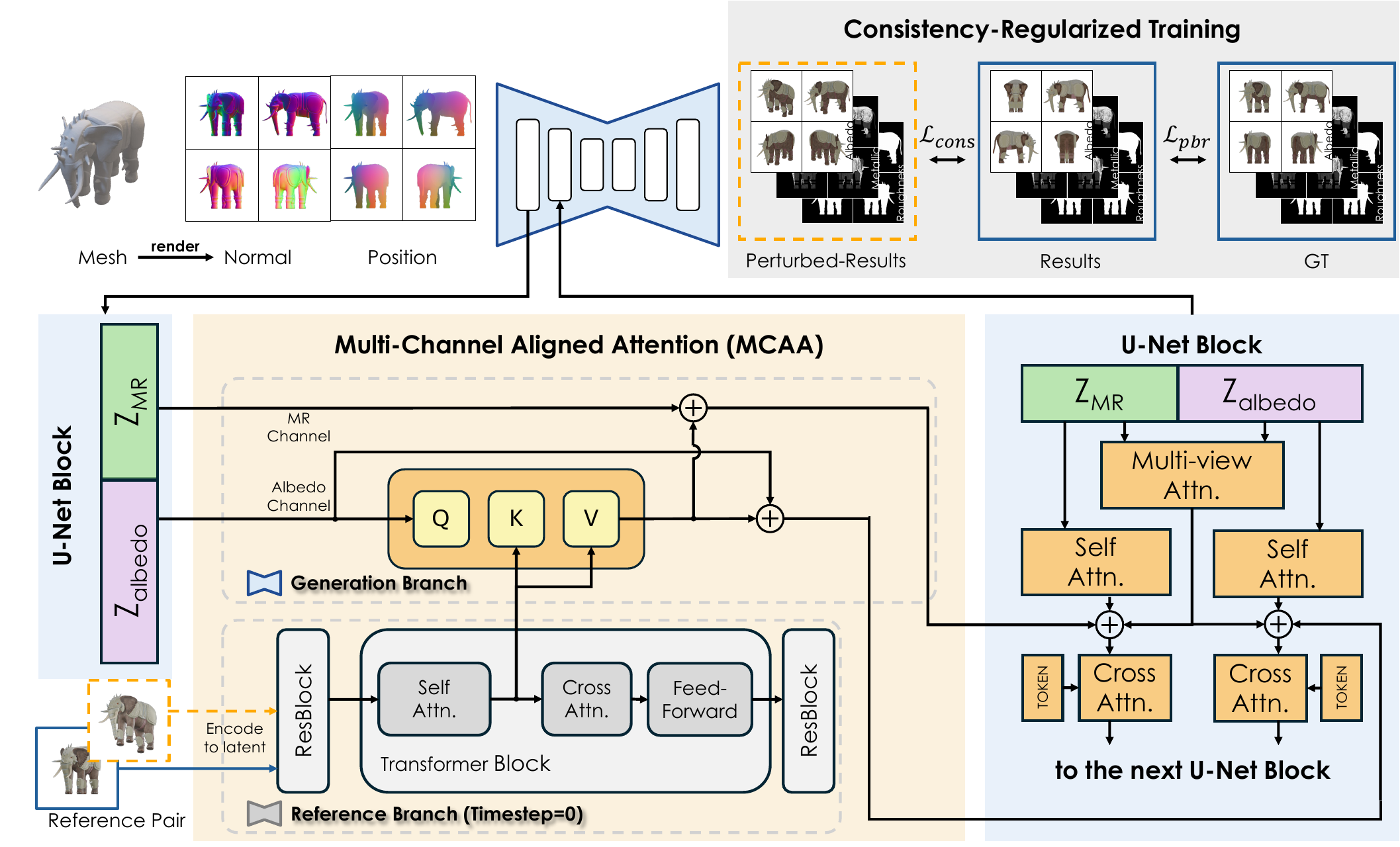}
    \vspace{-3.5mm}
    \caption{Overview of MaterialMVP, which takes a 3D mesh and a reference image as input and generates high-quality PBR textures through multi-view PBR diffusion. The albedo and MR channels are aligned via MCAA, effectively reducing artifacts. And Consistency-Regularized Training effectively eliminates illumination and addresses material inconsistency across multi-view PBR textures. }
    \label{fig:pipeline}
    \vspace{-3.5mm}
\end{figure*}

\noindent\textbf{Texture Generation.}
The development of 3D generation has significantly advanced the creation of high-quality textures for 3D meshes \cite{yu2023texture,richardson2023texture,cao2023texfusion,chen2023text2tex,le2023euclidreamer,tang2024intex,chen2024scenetex,zeng2024paint3d,zhang2024repaint123,bensadoun2024meta,lu2024direct2,zhao2025hunyuan3d,xiang2024make,yu2024texgen}. Extending text-to-image diffusion models\cite{ho2020denoising,song2020denoising,rombach2022high,podell2023sdxl,li2024hunyuan} provides a foundation for generating detailed and semantically coherent textures from textual descriptions \cite{richardson2023texture, cao2023texfusion, chen2023text2tex, le2023euclidreamer, tang2024intex, chen2024scenetex, liu2024text, zhang2024texpainter, bensadoun2024meta, lu2024direct2}. Building upon this, using image prompts as input enables textures to closely align with specific visual references \cite{zeng2024paint3d,
zhang2024repaint123, zhao2025hunyuan3d, zhang2024clay,yu2024texgen}. Advancing further, 3D priors, such as depth maps \cite{le2023euclidreamer, tang2024intex, zhang2024repaint123, xiang2024make}, normal maps \cite{bensadoun2024meta, lu2024direct2, zhao2025hunyuan3d}, and position maps \cite{bensadoun2024meta, zhao2025hunyuan3d, zeng2024paint3d}, ensures that the textures are not only visually realistic but also geometrically consistent with the underlying 3D structure.

\noindent\textbf{PBR Generation.}
PBR simulates light-surface interactions with physically accurate material properties, driving extensive research into PBR information estimation \cite{chen2023fantasia3d, zhang2024dreammat, wu2023hyperdreamer, xu2023matlaber, yeh2024texturedreamer, vainer2024collaborative, sartor2023matfusion, vecchio2024matfuse, xiang2024make, youwang2024paint, liu2024unidream, chen2024intrinsicanything, zhang2024mapa, fang2024make, xiong2024texgaussian} for high-quality 3D texture generation. Generation-based approaches \cite{vainer2024collaborative, sartor2023matfusion, vecchio2024matfuse, chen2024intrinsicanything, zeng2024rgb} leverage diffusion models to learn material priors and recover PBR properties through physical rendering; retrieval-based techniques \cite{zhang2024mapa, fang2024make} adapt pre-built material graphs from libraries to ensure visual consistency and editability; optimization-based methods \cite{chen2023fantasia3d, zhang2024dreammat, wu2023hyperdreamer, xu2023matlaber, yeh2024texturedreamer, youwang2024paint, liu2024unidream} first generate initial textures and then refine them through techniques like Score-Distillation Sampling \cite{poole2022dreamfusion}.

\noindent\textbf{Multi-view Generation.}
Multi-view generation \cite{wang2023imagedream, liu2024oneA, liu2024oneB, hu2024mvd, ding2024text, tang2024mvdiffusion++, wen2024ouroboros3d, woo2024harmonyview, long2024wonder3d, yang2024hunyuan3d, shi2023zero123++, shi2023mvdream, li2024era3d, li2025multi, li2025cmd, li2024pshuman} has been adopted in texture generation \cite{liu2024text, zhang2024texpainter, bensadoun2024meta, lu2024direct2, zhao2025hunyuan3d, deng2024flashtex, zhang2024clay} to address issues such as blurring or artifacts.
Some methods integrate viewpoint information or 3D priors \cite{chen2024cascade, jeong2024nvs, liu2023syncdreamer, yang2024consistnet, yang2024viewfusion, hollein2024viewdiff}. 
Other approaches utilize epipolar geometry to improve consistency between views or reduce memory overhead \cite{li2024era3d, huang2024epidiff, kant2024spad}, while another technique tiles multiple view images into a single layer, treating them as a unified input during the diffusion model’s denoising process \cite{shi2023zero123++, shi2023mvdream, wu2024unique3d, li2023instant3d}.

\section{Method}
\label{sec:method}

As shown in~\cref{fig:pipeline}, our MaterialMVP generates PBR textures conditioned on a 3D mesh and an image prompt. The 3D mesh is input in the form of normal maps and position maps, which are encoded into latent space and concatenated channel-wise with noise latent as U-Net input. The generated albedo map is expected to be free from lighting information, while the metallic and roughness maps should be both accurate and precisely aligned. In this section, we first review the multi-view image generation diffusion model and the principles of PBR-based material modeling, which form the foundation of our approach. Then, we introduce the Consistency-Regularized Training method, designed to disentangle residual lighting information from the albedo map while improving robustness against input variations. Finally, we present our dual-channel material generation framework, which ensures the precise alignment and fidelity of the generated texture maps.


\subsection{Preliminary}
\label{subsec:pre}
\noindent\textbf{Multi-view Diffusion.}
The Latent Diffusion Model (LDM)~\cite{rombach2022high} is a generative framework that operates in a compressed latent space, combining the principles of diffusion processes with VAEs to generate high-quality images efficiently. 
Building upon LDM, existing methods~\cite{long2024wonder3d, wang2023imagedream, tang2024mvdiffusion++} propose multi-view diffusion models that extends the latent space $z$ to multi-view representations $Z=\left \{z_{1},\dots ,z_{n} \right \}$ via multi-view attention as
\begin{equation}
    z_i^{new} = {\sum_{j=1}^{n}} \text{Softmax}(\frac{ Q_i K_j^T}{\sqrt{d}}) \cdot V_j,
    \label{eq:mvattn}
\end{equation}
where $Q$, $K$, and $V$ are the projected features of Query, Key, and Value, enabling synchronized denoising of geometrically consistent multi-view outputs.

\noindent\textbf{PBR Material.}
Our material representation employs the Disney Principled Bidirectional Reflectance Distribution Function (BRDF) framework~\cite{burley2012physically}, defining surfaces through three parameters: albedo, metallic, and roughness. These parameters are stored in two separate textures: a combined MR map storing metallic and roughness data, while albedo remains in an RGB texture map.

\subsection{Consistency-Regularized Training}
\label{subsec:CRTrain}
We observe two key limitations in multi-view PBR synthesis with image prompts: (1) \textbf{view sensitivity}, where slight perturbations in camera pose can lead to dramatically different material outputs, and (2) \textbf{illumination entanglement}, where lighting from the reference images is incorrectly baked into the output, or the pretrained diffusion model generates unintended lighting effects.

To address these issues, we propose a consistency-regularized training strategy that trains the model on pairs of reference images rather than individual samples. At each training step, the diffusion model is jointly conditioned on two reference images. This dual-prompt mechanism introduces implicit geometric consistency constraints to stabilize multi-view generation while encouraging the model to disentangle lighting variations from material features. 

\begin{figure}[t]
    \centering
    \includegraphics[width=0.8\linewidth]{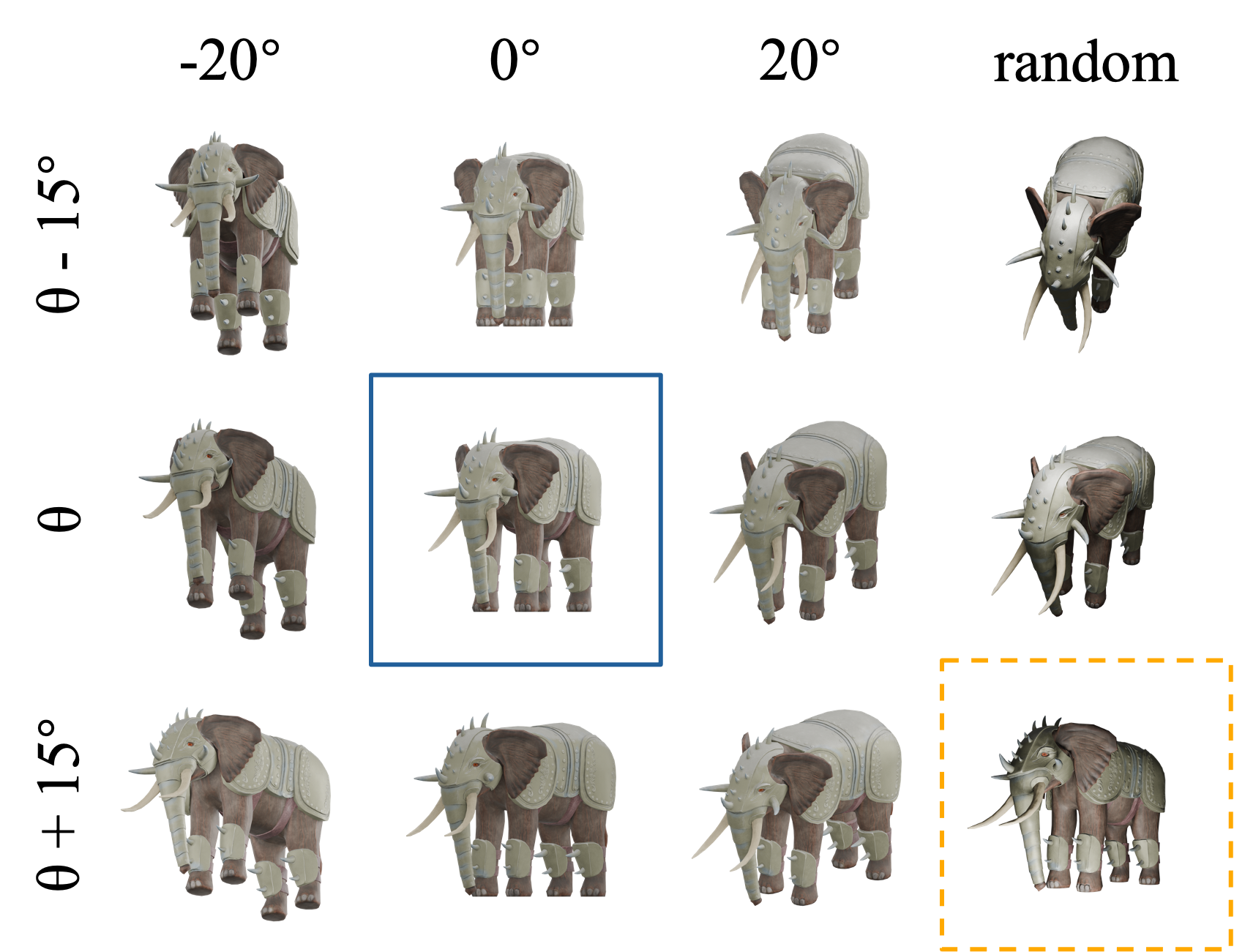}
    \vspace{-3.5mm}
    \caption{\textbf{Reference Image Pair Selection.} The selected image pair exhibits perturbations in camera pose and/or lighting. The solid box indicates the chosen reference image, and the dashed box indicates the perturbed reference image.}
    \label{fig:img_pair}
    \vspace{-5.5mm}
\end{figure}

\subsubsection{Reference Pair Selection}
\label{subsubsec: RefPair}

For each object, we construct a candidate set $\mathcal{I}$ of 312 images rendered in various camera poses and lighting conditions at elevation angles $\theta_{c} \in\left\{-20^{\circ}, 0^{\circ}, 20^{\circ}, \text { random}\right\}$, with 24 azimuth angles $\phi_{c}$ uniformly sampled per elevation. For $\theta_{c} \in\left\{-20^{\circ}, 0^{\circ}, 20^{\circ}\right\}$, each pose is rendered under three random HDR lighting conditions; for $\theta_{c}=\text{random}$, each pose is rendered under a single point light with random light elevation $\theta_{l} \in\left\{-30^{\circ}, 0^{\circ}, 30^{\circ}\right\}$ and light azimuth $\phi_{l} \in\left\{-45^{\circ}, 0^{\circ}, 45^{\circ}\right\}$.
As shown in \cref{fig:img_pair}, to select a reference pair $(I_1,I_2)$, we first choose $I_1$ from $\mathcal{I}$ with probability $p$ favoring point light images (in practice, we set $p=0.4$), and then select $I_2$ from the subset $\mathcal{S}(I_1)$ defined as
\begin{equation}
\mathcal{S}\left(I_{1}\right)=\left\{I \in \mathcal{I} \mid \phi_{c}(I) \in\left\{\phi_{c}\left(I_{1}\right), \phi_{c}\left(I_{1}\right)\pm15^{\circ}\right\}\right\},\nonumber
\end{equation}
where $\phi_{c}(I)$ denotes the camera azimuth angle of image $I$. The selection process can be compactly expressed as
\begin{equation}
(I_1, I_2) = \left( \text{Sample}(\mathcal{I}, p), \text{Sample}(\mathcal{S}(I_1)) \right),\nonumber
\end{equation}
where $\text{Sample}(\mathcal{I}, p)$ samples $I_1$ with probability $p$ for point light images, and $\text{Sample}(\mathcal{S}(I_1))$ uniformly selects $I_2$ from $\mathcal{S}(I_1)$. In this way, we obtain a reference image pair with subtle differences in viewpoint and lighting.

\begin{table*}[t]
\centering
\caption{Quantitative comparison with state-of-the-art methods. We compare with two classes of methods, one conditioned on text only, and the other one based on image. Our method achieves the best performance compared with both classes.}
\vspace{-3.5mm}
\setlength{\tabcolsep}{5pt}\small
\begin{tabular}{ccccccc}
\toprule 
Method   & Condition     & {CLIP-FID$\downarrow$} & {FID$\downarrow$} & {CMMD$\downarrow$} & {CLIP-I$\uparrow$} & {LPIPS$\downarrow$}  \\ \midrule
Text2Tex \cite{chen2023text2tex} \textcolor{blue}{\textsubscript{ICCV'23}}  & Text          & 31.83    & 187.7      & 2.738 & -      & 0.1448 \\
SyncMVD \cite{liu2024text} \textcolor{blue}{\textsubscript{SIGGRAPH Asia'24}}   & Text          & 29.93    & 189.2      & 2.584 & -      & 0.1411 \\
Paint-it \cite{youwang2024paint} \textcolor{blue}{\textsubscript{CVPR'24}} & Text          & 33.54    & 179.1      & 2.629 & -      & 0.1538       \\
Paint3D \cite{zeng2024paint3d} \textcolor{blue}{\textsubscript{CVPR'24}} & Text          & 30.17    & 185.7      & 2.755 & -      & 0.1388 \\
\hline
Paint3D \cite{zeng2024paint3d} \textcolor{blue}{\textsubscript{CVPR'24}}& Image          & 26.86    & 176.9      & 2.400 & 0.8871      & 0.1261       \\
TexGen \cite{yu2024texgen} \textcolor{blue}{\textsubscript{TOG'24}}  & Text + Image & 28.23    & 178.6      & 2.447 & 0.8818 & 0.1331       \\ 
Ours     & Image         & \textbf{24.78}    & \textbf{168.5}      & \textbf{2.191} & \textbf{0.9207} & \textbf{0.1211}       \\ \bottomrule
\end{tabular}
\label{tab: comparisons}
\vspace{-5.5mm}
\end{table*}

\subsubsection{Training Strategy}

Following the input of reference image pair $(I_1, I_2)$, the network processes each image sequentially while enforcing output equivalence through consistency regularization. Specifically, at each training timestep $t$ the model $\epsilon_\theta$ generates diffusion noises $\epsilon_t^1$ and $\epsilon_t^2$ by conditioning on $I_1$ and $I_2$ respectively. We arbitrarily select one of them to compute the PBR loss, using $\epsilon_t^1$ as an example
\begin{equation}
\epsilon_t^1 = \epsilon_\theta(z_t,t,c(I_1)),\nonumber \ \ \ 
\epsilon_t^2 = \epsilon_\theta(z_t,t,c(I_2)),\nonumber
\end{equation}
\begin{equation}
\mathcal{L}_{\text{pbr}} = \mathbb{E}_{\epsilon \sim \mathcal{N} (0,1),t}[\|\epsilon - \epsilon_t^1\|_2^2].\nonumber
\end{equation}
To achieve consistency under varying camera poses and lighting conditions, we enforce $\epsilon_t^1 \approx \epsilon_t^2$. This is formulated through an $L_2$ consistency loss as
\begin{equation}
\mathcal{L}_{\text{cons}} = \mathbb{E}_{t}[ \|\epsilon_t^1 - \epsilon_t^2\|_2^2],\nonumber
\end{equation}
which is combined with the PBR loss via a weighting parameter $\lambda=0.1$ as
\begin{equation}
    \mathcal{L} = (1-\lambda)\mathcal{L}_{\text{pbr}} + \lambda \mathcal{L}_{\text{cons}}.
    \label{eq:loss}
\end{equation}

\subsection{Dual-Channel Material Generation}
\label{subsec:dual_channel}

Albedo and MR textures serve distinct roles in PBR material synthesis. However, existing approaches are not designed to handle the unique requirements of PBR material generation, particularly the need to address the inherent differences between albedo and MR. To bridge this gap, we propose strategies that independently optimize albedo and MR generation, ensuring not only precise alignment but also high-quality texture synthesis.

\subsubsection{Multi-Channel Aligned Attention}
Some approaches guide diffusion generation by incorporating conditional image prompts through a reference diffusion branch, where the latent representation $z_{ref}$ is injected into the generation branch via cross-attention. However, when applied to PBR material generation, it is unsuitable due to the distributional gap between the reference image and the MR texture. Specifically, attention mechanisms struggle to align these mismatched latent features, leading to inconsistent material synthesis and texture degradation, particularly in geometrically flat regions of 3D meshes where subtle material variations are critical.

To resolve these issues, as illustrated in \cref{fig:pipeline}, we propose Multi-Channel Aligned Attention that independently optimizes the albedo and MR channels while maintaining their material consistency. Specifically, the albedo channel retains the conventional reference-guided cross-attention 
\begin{equation}
    \text{Attn}_{albedo} = \text{Softmax}\left(\frac{ Q_{albedo} K_{ref}^T}{\sqrt{d}}\right) \cdot V_{ref},
    \label{eq:mcaa_albedo}
\end{equation}
ensuring that the generated albedo correctly follows the reference image.
In contrast, the MR latent $z_{MR}$ discards direct reference conditioning and instead leverages residual connections from the albedo channel
\begin{equation}
    z_{MR}^{new} = z_{MR} + \text{Attn}_{albedo},
    \label{eq:mcaa_mr}
\end{equation}
allowing it to inherit spatial coherence and geometric priors from the albedo features.

This design addresses the unalignment between different domains of PBR material, and maintains generation diversity without increasing parameter complexity, as no additional trainable parameters are introduced beyond the baseline cross-attention framework.

\subsubsection{Learnable Material Embeddings}
Inspired by IC-Light \cite{zhang2025scaling}, we introduce learnable embeddings for both the albedo and MR channels to explicitly model their differences. Specifically, we initialize two independent 16×1024 embeddings, which are injected into the respective channels via cross-attention layers as shown in \cref{fig:pipeline}. Both the embeddings and the attention modules are trainable, enabling the model to effectively capture the distinct characteristics of albedo and MR textures.
\section{Experiment}
\label{sec:experiment}

\begin{figure*}[t]
    \centering
    \includegraphics[width=0.85\linewidth]{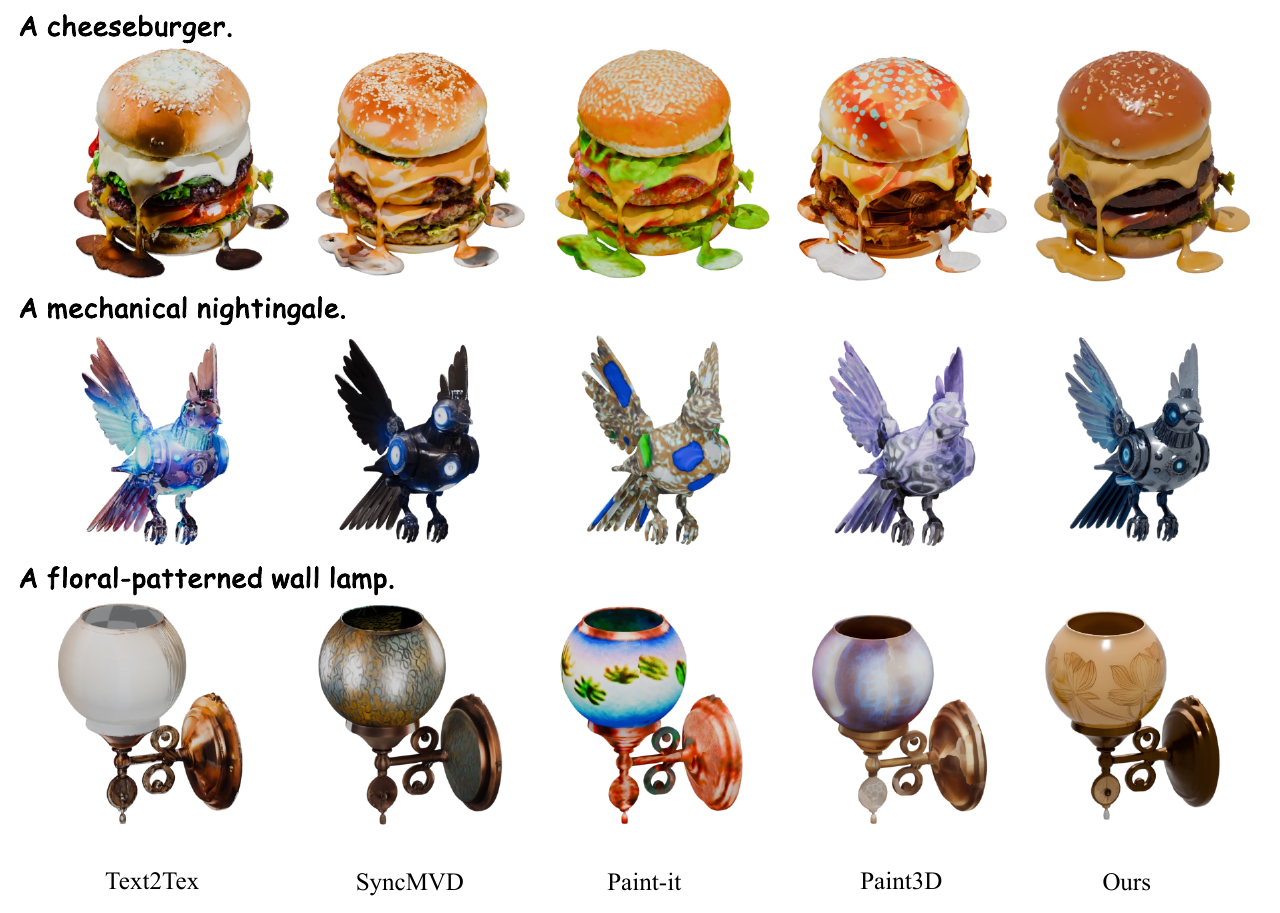}
    \vspace{-3.5mm}
    \caption{Qualitative comparison of text-conditioned 3D generation methods and our approach. Our method achieves superior fidelity in texture and pattern coherence, avoiding blur, inconsistent lighting, and misaligned patterns found in other approaches.}
    \label{fig: t_comparison}
    \vspace{-3mm}
\end{figure*}

\noindent\textbf{Dataset.}
Our training dataset consists of 70,000 high-quality 3D assets selected from Objaverse \cite{deitke2023objaverse} and Objaverse-XL \cite{deitke2023objaversexl}. For each 3D object, we rendered data from four elevation angles: $-20^{\circ}$, $0^{\circ}$, $20^{\circ}$, and a randomly sampled angle. At each elevation, we captured 24 uniformly distributed views, generating corresponding albedo, metallic, roughness maps, and HDR/Point-light images of 512 $\times$ 512 resolution. During each training step, we selected 6 sets of uniformly distributed PBR maps from the same elevation and 2 HDR/Point-light images as reference data (see \cref{subsubsec: RefPair} for the selection methodology). 

\noindent\textbf{Evaluation Metrics.}
We use Fréchet Inception Distance (FID), CLIP-based FID (CLIP-FID), and Learned Perceptual Image Patch Similarity (LPIPS) to measure the similarity between the generated textures and the ground truth. CLIP Maximum-Mean Discrepancy (CMMD) is used to assess the diversity and richness of the generated texture details. And CLIP-Image Similarity (CLIP-I) is employed to evaluate how well the generated textures semantically align with the input images (for image prompt methods).

\noindent\textbf{Experimental Setup.}
Our model is initialized from the ZSNR checkpoint \cite{lin2024common} of Stable Diffusion 2.1 and trained using the AdamW optimizer with a learning rate of $5 \times 10^{-5}$. The training process includes 2000 warm-up steps and requires approximately 180 GPU days.

\begin{figure*}[t]
    \centering
    \includegraphics[width=0.85\linewidth]{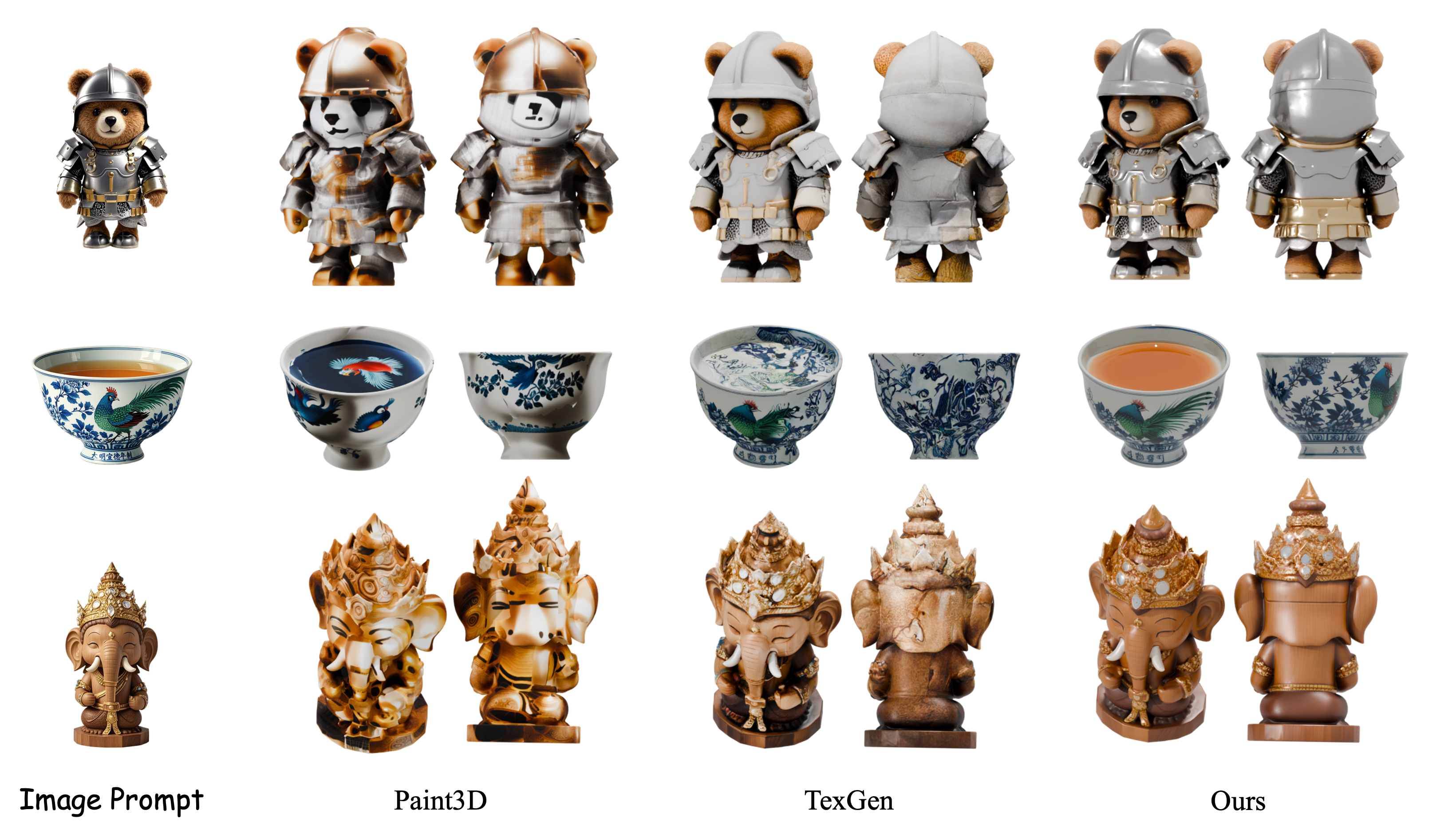}
   \vspace{-3.5mm}
    \caption{Qualitative comparison of image-conditioned 3D texture generation methods. It demonstrates that our method outperforms other methods in terms of faithfully following the image and its semantics, as well as generating textures with superior coherence.}
    \label{fig: i_comparison}
    \vspace{-5.5mm}
\end{figure*}

\begin{figure*}[t]
    \centering
    \includegraphics[width=0.9\linewidth]{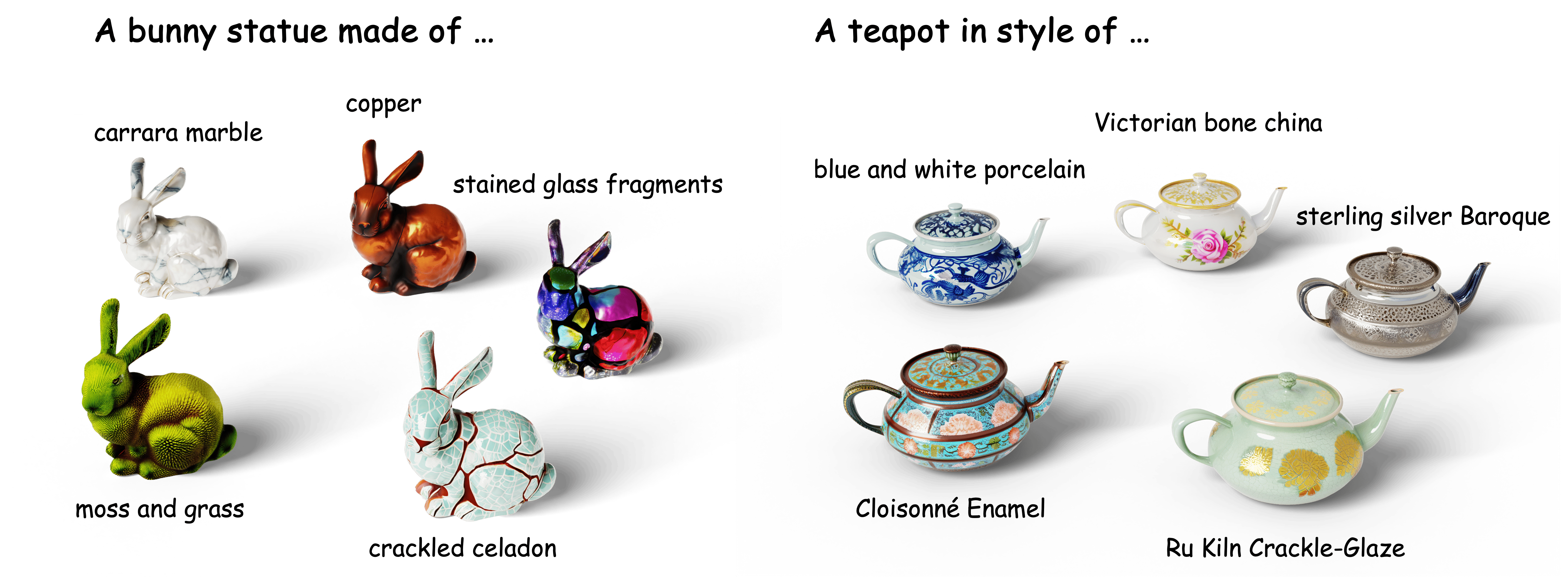}
    \vspace{-3.5mm}
    \caption{We show a variety of PBR textures for two different 3D assets, demonstrating the diversity and semantic alignment capabilities of our model. (Better viewed by zooming in)}
    \label{fig: re-texture}
    \vspace{-3.5mm}
\end{figure*}

\noindent\textbf{Compared Methods.}
We compare with text- and image-conditioned methods, including
 Text2Tex \cite{chen2023text2tex}, Paint3D \cite{zeng2024paint3d}, Paint-it \cite{youwang2024paint}, SyncMVD \cite{liu2024text}, and TexGen \cite{yu2024texgen}.

\subsection{Quantitative Comparison}
We at first conduct a quantitative evaluation to assess the performance of our proposed method, focusing on its ability to generate high-quality textures and maintain image alignment capabilities. We utilize a subset of Objaverse containing 176 objects as the evaluation set. The evaluation results are presented in \cref{tab: comparisons}, showcasing that our method demonstrates superior texture generation quality and semantic alignment capabilities.
As the table reports, the leading performance in FID and LPIPS metrics indicates a high similarity between our model's outputs and the ground truth images, demonstrating our model's ability to effectively follow the input reference images. The superior performance in the CMMD metric suggests that the textures generated by our model are rich in details. Additionally, the leading scores in CLIP-I and CLIP-FID metrics reveal that the generated textures are semantically well-aligned with the reference images, further validating the semantic consistency of our model's outputs.

\begin{figure*}[t]
    \centering
    \includegraphics[width=0.9\linewidth]{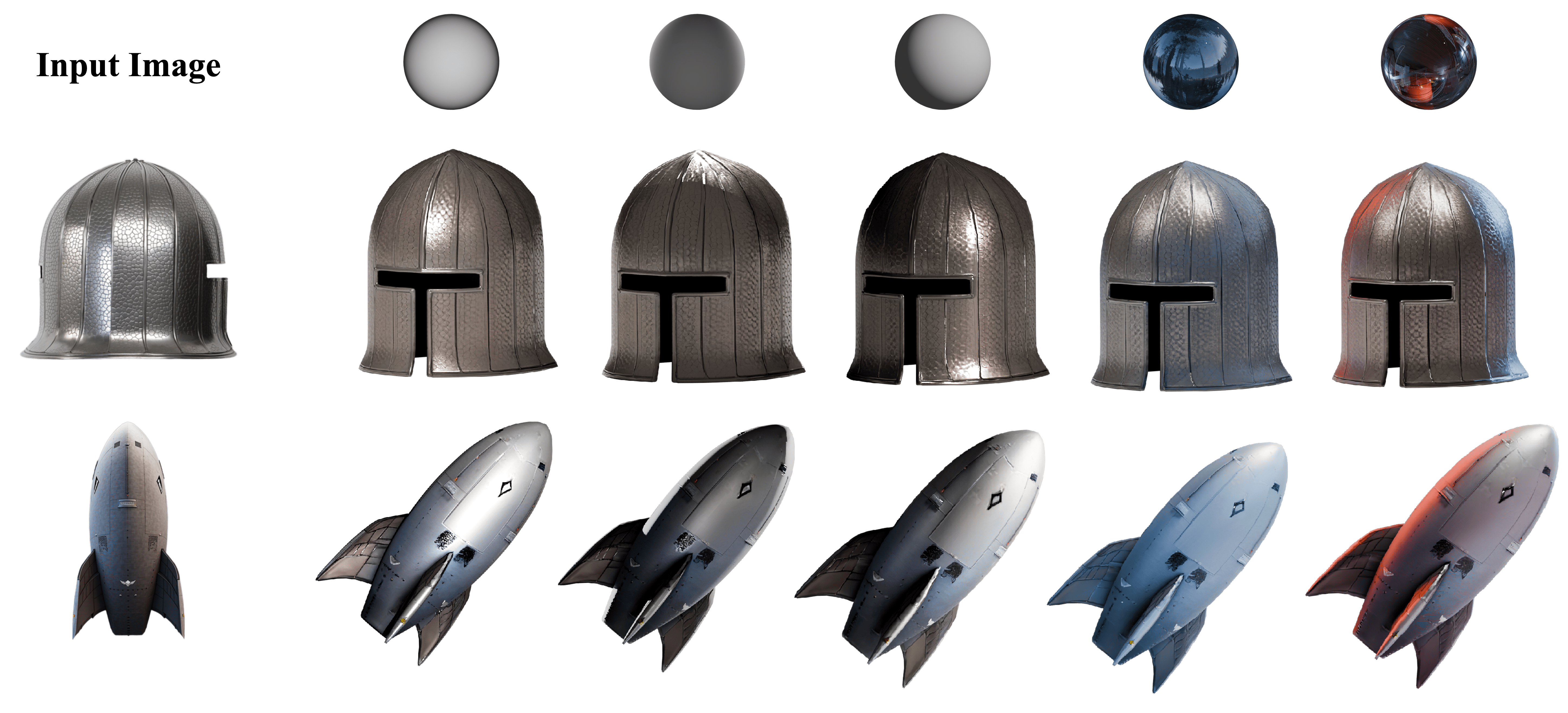}
    \vspace{-3.5mm}
    \caption{We show PBR textures for two meshes with image prompts and display them under various environmental illuminations. The results demonstrate high accuracy in our PBR textures and the ability to remain unaffected by the input image's lighting.}
    \label{fig: re-light}
    \vspace{-2.5mm}
\end{figure*}

\subsection{Qualitative Comparison}
For image-conditioned models, we use JimengAI \cite{JM_AI} to generate reference images as prompts. Qualitative results are presented in \cref{fig: t_comparison} and \cref{fig: i_comparison}. Our model demonstrates superior performance compared to other methods in terms of: 1) \textbf{Realism}. Our model exhibits astonishingly realistic effect, especially under various environmental illumination. Also, the consistency between multi-view PBR materials effectively avoids artifacts such as blurring, texture seams, and the Janus Effect. 2) \textbf{Illumination invariance}. Results in \cref{fig: i_comparison} show that our method has successfully eliminated the illumination present in the input image while accurately reflecting the environmental lighting through the generated PBR materials. In contrast, other methods presented in \cref{fig: t_comparison} and \cref{fig: i_comparison}, although capable of generating some illumination effects, fail to adhere to physical principles, resulting in visually unappealing outcomes. 3) \textbf{Global consistency}. Some methods achieve accurate alignment with the reference image on the front view (see \cref{fig: i_comparison}), while exhibiting poor consistency on the back views. This inconsistency manifests itself as chaotic patterns, unwanted shadows, or even illogical textures. The inconsistency is also observed in \cref{fig: t_comparison}. In contrast, our method generates seamless and consistent textures, reasonably extending the content beyond the reference image and maintaining global coherence. Furthermore, our method achieves precise alignment with the input image both at the pixel level and semantically.

\begin{figure*}[t]
    \centering
    \includegraphics[width=0.9\linewidth]{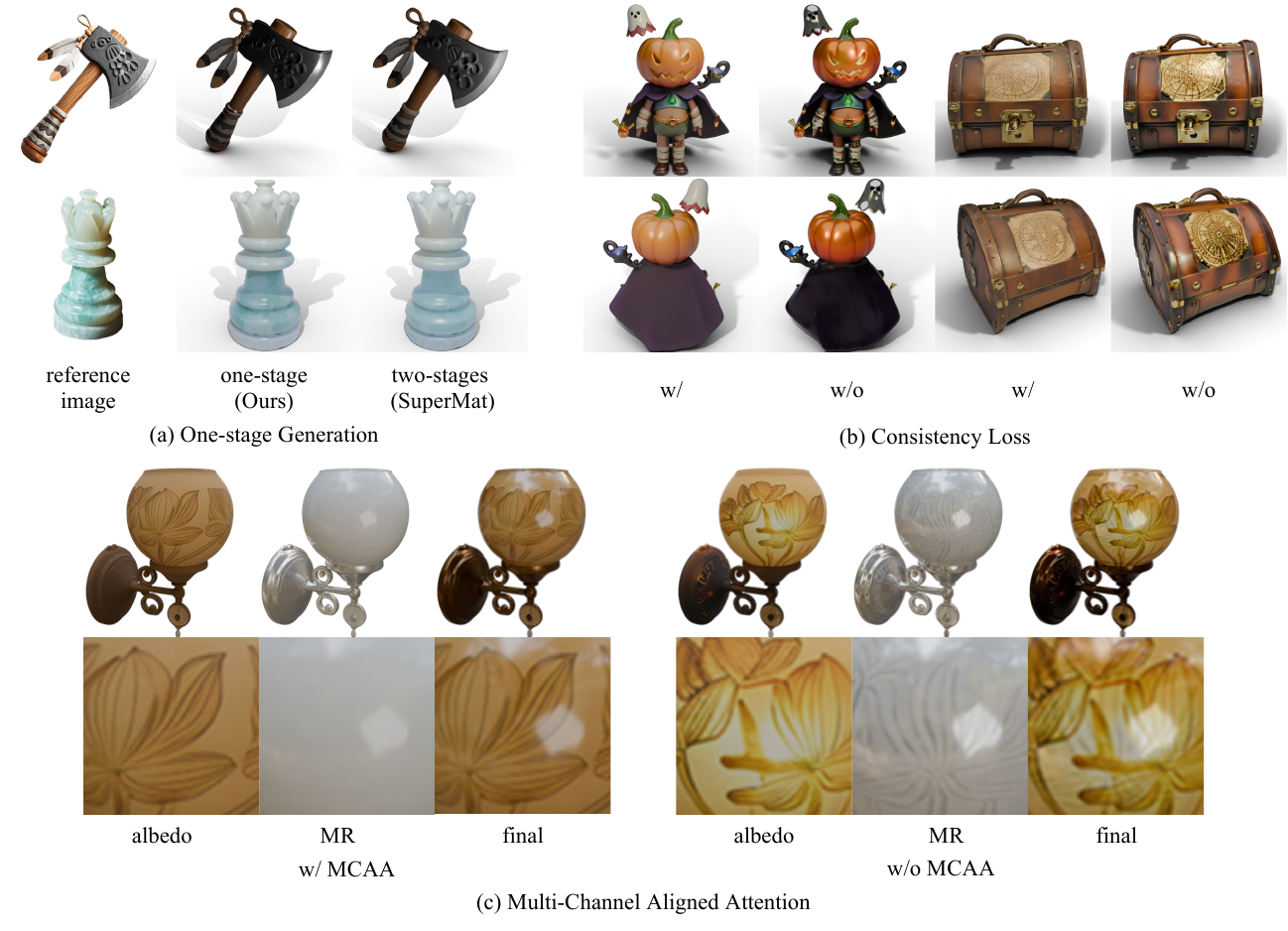}
    \vspace{-3.5mm}
    \caption{We ablate the one-stage generation scheme, the consistency loss, and the MCAA, validating the effectiveness of our method in generating high-quality and semantically accurate textures, as well as eliminating blurring artifacts caused by PBR misalignment.}
    \label{fig: ablation}
    \vspace{-5.5mm}
\end{figure*}

\subsection{Ablation Study}

\noindent\textbf{One Stage Generation Strategy.}
To demonstrate the superiority of the end-to-end generation of multi-view PBR maps over the two-stage approach, which first generates multi-view images and subsequently converts them into PBR maps, we replicate a representative work of this method, SuperMat \cite{hong2024supermat}. It can be observed in \cref{fig: ablation} (a) that SuperMat is not able to accurately estimate materials, often rendering glass or metal with a plastic-like appearance, which is likely due to the accumulated errors during the process of generating multi-view images from a single RGB image. Therefore, the end-to-end generation method is better suited to handle this task.

\noindent\textbf{Consistency Loss.}
To show the importance of the consistency loss in improving PBR map accuracy during training, we set the weighting parameter  $\lambda$ to zero and trained the model till convergence. The results in \cref{fig: ablation} (b) indicate that, without the consistency loss, the model frequently mispredicts materials with excessively high metallic properties, assigning a metallic appearance to various materials.

\noindent\textbf{MCAA.}
To verify the effectiveness of MCAA, we train a variant where MCAA is replaced with standard weight sharing between dual channels. As \cref{fig: ablation} (c) shows, we predict the albedo and MR maps for the same mesh, and subsequently generate the final textured mesh. The variant without MCAA produces misaligned albedo and MR maps in detailed regions, ultimately leading to blurred textures. It is worth noting that generating an MR map that is perfectly aligned with the albedo or generating a blank MR map can both achieve artifact-free results in the final output. 

\subsection{Application}
Our model shows promising capability in generating high-quality textures. As illustrated in \cref{fig: re-texture}, it enables the creation of textures with realistic material properties using different prompts. 
Moreover, as shown in \cref{fig: re-light}, we generate two objects and render them under five distinct environmental illuminations. The textures generated by our model exhibit exceptional fidelity and physical accuracy under various lighting conditions.

\section{Conclusion}
\label{sec:conclusion}

We present MaterialMVP, a novel one-stage framework that generates high-quality PBR textures for 3D meshes from image prompts. We extend the conventional multi-view diffusion to a multi-view PBR diffusion process, generating consistent albedo, metallic, and roughness maps across multiple views. We also incorporate an additional perturbed reference image during training, ensuring stable texture generation in varying lighting conditions or unstable viewpoints. Experiments demonstrate MaterialMVP's superior performance in both text-to-texture and image-to-texture generation tasks.

\section*{Acknowledgment}
This work was supported in part by the National Natural Science Foundation of China (Grant No. 62372480), in part by the Guangdong Basic and Applied Basic Research Foundation (No. 2023A1515012839), in part by 2025 Tencent AI Lab Rhino-Bird Focused Research Program, in part by Huawei (No. HUAWEI25IS02 and No. 25260020L082), in part by gift fund from Wiener Intelligence (No. HKWITL25IS01), and in part by HKUST-MetaX Joint Lab Fund (No. METAX24EG01-D).

{
    \small
    \bibliographystyle{ieeenat_fullname}
    \bibliography{main}
}

\end{document}